# Geometric Jacobians Derivation and Kinematic Singularity Analysis for Smokie Robot Manipulator & the Barrett WAM

Reza Yazdanpanah A.

*Abstract*— This paper discusses deriving geometric jacobians and identifying and analyzing the kinematic singularities for two 6 DOF arm robots. First we show the direct kinematics and D-H parameters derived for these two arms. The Geometric Jacobian is computed for Barrett WAM and Smokie OUR. By analyzing the Jacobian matrices we find the configurations at which J is rank-deficient and derive the kinematic singularities through J's determinent. Schematic are provided to show the singular configurations of both robots. Finally a survey is done on redundant kinematic allocation schemes for 7 DoF Barrett WAM.

*Index Terms*—Geometric Jacobian, Kinematic Singularity, Smokie Robot, Barrett WAM, Redundant kinematic allocation

## I. INTRODUCTION

THIS paper discusses the Geometric Jacobians and Kinematic Singularity derivation for 6 DoF version of Barrett WAM and Smokie OUR arm. The direct kinematics and modelling of both arms are done on previous project and result are used here. By having The D-H parameters, we follow the Jacobian computation procedure [1] by which a systematic, general method is used to derive the Jacobian matrix.

Jacobian matrices analysis reveals that they are not full rank matrices. So, there are configurations at which Jacobians are rank-deficient. These configurations are named as Kinematic Singularities.

Avoiding singularities is an important topic in robot manipulation. At singularities, the mobility of the structure is reduced, therefore the arbitrary motion of end effector is not possible anymore. The problem arises here is that the inverse kinematics solution may has infinite solutions. So the singularity configurations must be avoided.

In this paper first we shedlight on Smokie robot and WAM arm and choose the joints and links. Then the D-H parameters are derived and joint limits are specified. Next the Jacobian matrices are computed and singularity points are derived through their determinants. A discussion is done on singularity configurations and the schematic plots are shown. Finally a literature survey is done on redundant kinematic allocation schemes for the Barrett WAM.

## II. MANIPULATORS

Manipulators are robots with a mechanical arm operating under computer control. They are composed of links that are connected by joints to form a kinematic chain. The manipulators that is considered in this paper have solely rotary, also called revolute, joints. Each represents the interconnection between two links. The axis of rotation of a revolute joint, denoted by $Z_i$, is the interconnection of links $l_i$ and $l_{i+1}$. The joint variables, denoted by $q_i$, represent the relative displacement between adjacent links

In this paper we study two Manipulators: the Smokie Robot and the Barrett Whole Arm Manipulation (WAM) arm. These two are introduced in this section.

### A. Smokie Robot OUR

The OUR is a low-cost, 6-DOF industrial manipulator manufactured by Smokie Robots. With a weight of 18.4 kg it is a lightweight manipulator. It has a reach of 85 cm and a maximal payload of 5 kg and is shown in Figure 1.

OUR is a very low cost robot that has the comparable performance with many general industrial robots. OUR's modularized design enables users to reconfigure the robot system with 4-7 DoFs to meet their requirements. The standard OUR is designed with 6-DoFs.

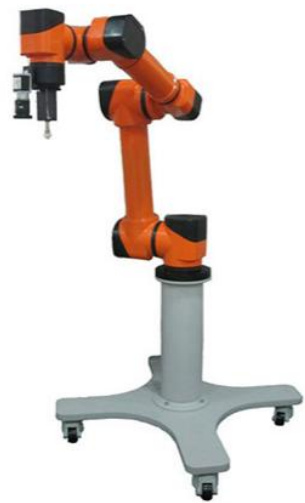

*Figure 1: Smokie OUR arm*

Reza Yazdanpanah Abdolmalaki is with The University of Tennessee Knoxville, TN 37996 USA (e-mail: ryazdanp@vols.utk.edu).

*1) Links and Joint Identification*

OUR Smokie robot consists of 6 links and 6 revolute joints. Each joint connects two consecutive links to each other. All the 6 revolute joints have $-180°{\sim}180°$ rotation capability. The links, Joints and Dimensions are shown in figure 2.

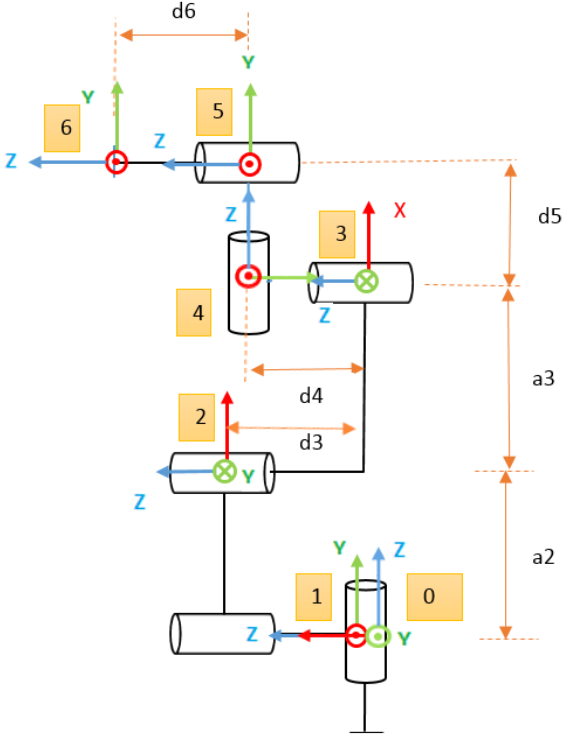

*Figure 2: OUR links and Joint Identification*

*2) D-H parameters*

The commonly used DH-convention defines four parameters that describe how the reference frame of each link is attached to the robot manipulator. Starting with the inertial reference frame, one additional reference frame is assigned for every link of the manipulator. The four parameters $a_i, d_i, \alpha_i, \theta_i$ defined for each link $i \in [1, n]$ transforms reference frame $i - 1$ to $i$ using the four basic transformations.

For this purpose we choose the coordinates as Figure 2. The algorithm presented in D-H convention is used to assign the proper coordinates for OUR and the parameters are shown in Table 1.

*Table 1: OUR D-H Table*

| $i$ | $a_i$ | $\alpha_i$ | $d_i$ | $\theta_i$ |
|---|---|---|---|---|
| 1 | 0 | $\pi/2$ | 0 | $\theta_1$ |
| 2 | 0.43 | 0 | 0.145 | $\theta_2$ |
| 3 | 0.336 | 0 | -0.145 | $\theta_3$ |
| 4 | 0 | $-\pi/2$ | 0.115 | $\theta_4$ |
| 5 | 0 | $\pi/2$ | 0.115 | $\theta_5$ |
| 6 | 0 | 0 | 0.115 | $\theta_6$ |

B. *WAM Arm*

The WAM Arm is a 7-degree-of-freedom (7-DOF) manipulator with human- like kinematics. With its aluminum frame and advanced cable-drive systems, including a patented cabled differential, the WAM is lightweight with no backlash, extremely low friction, and stiff transmissions. All of these characteristics contribute to its high bandwidth performance. The WAM Arm is the ideal platform for implementing Whole Arm Manipulation (WAM), advanced force control techniques, and high precision trajectory control. WAM is shown in figure 3.

The WAM Arm is a highly dexterous backdrivable manipulator. It is the only commercially available robotic arm with direct-drive capability between the motors and joints, so its joint-torque control is unmatched and guaranteed stable.

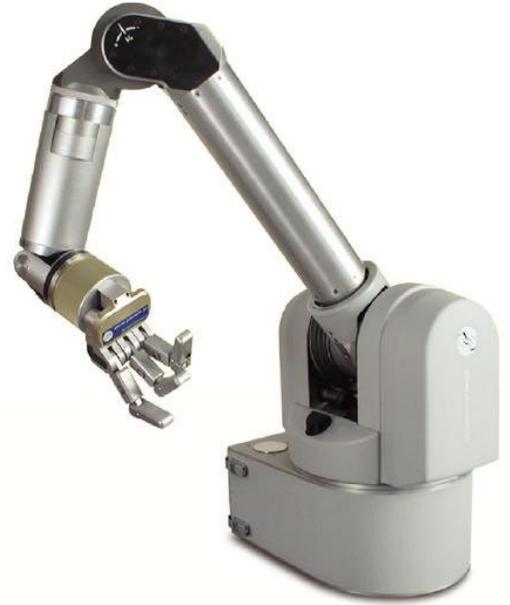

*Figure 3: BARRETT WAM arm*

*1) Links and Joints Identification*

In this section we identify the links and joints of each Arm and designate the degrees of freedom of both manipulators.

WAM Arm robot has 7 DoFs. It has 7 links and 7 joints. In this study, by assuming the lower arm rotational joint as fixed we consider a 6DOF version of the WAM. The Joints and dimensions are shown in Figure 4

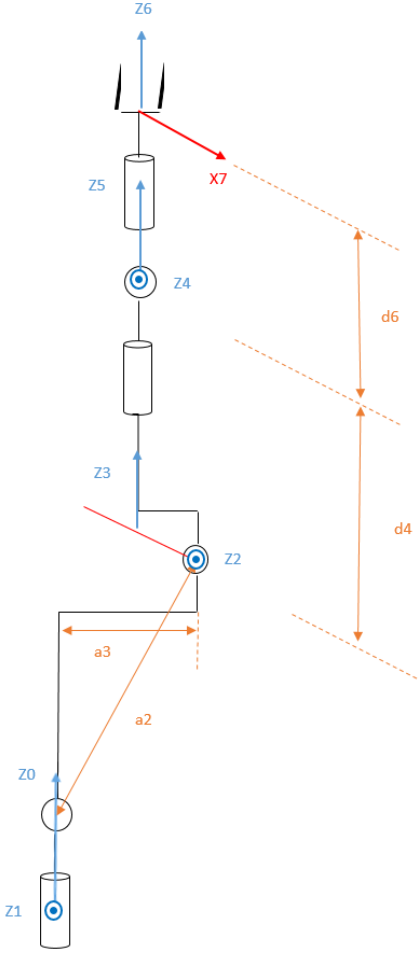

*Figure 4: Links, Joints and Axis of 6 DoF WAM*

### 2) D-H parameters of WAM

Exactly the same as previous procedure, we will first assign the proper coordinates for WAM based on D-H convention.

By assigning the coordinates, The D-H parameters can be calculated and are in the Table 2.

*Table 2: WAM D-H Table Parameters*

| i | $a_i$ | $\alpha_i$ | $d_i$ | $\theta_i$ |
|---|---|---|---|---|
| 1 | 0 | $-\pi/2$ | 0 | $\theta_1$ |
| 2 | $\sqrt{0.55^2 + 0.045^2}$ | 0 | 0 | $\theta_2$ |
| 3 | -0.045 | $\pi/2$ | 0 | $\theta_3$ |
| 4 | 0 | $-\pi/2$ | 0.3 | $\theta_4$ |
| 5 | 0 | $\pi/2$ | 0 | $\theta_5$ |
| 6 | 0 | 0 | 0.06 | $\theta_6$ |

## III. GEOMETRIC JACOBIAN

In the previous paper, The Direct Kinematics of manipulators were derived. The direct kinematic function for arms is expressed by homogeneous transformation matrix ($T_e^b(q)$) which describes the position and orientation of end effector with respect to reference base.

$$T_e^b(q) = \begin{bmatrix} n_e^b(q) & s_e^b(q) & a_e^b(q) & p_e^b(q) \\ 0 & 0 & 0 & 1 \end{bmatrix}$$

Where $q$ is the (n x 1) vector of joint variables $n_e, s_e, a_e$ are the unit vectors of a frame attached to the end effector, and $p_e$ is the position vector of the origin of such frame with respect to the origin of the base frame. $n_e, s_e, a_e$ and $p_e$ are all a function of $q$.

In this paper we want to discuss about "Differential Kinematics". The goal of differential kinematics is to find the relationship between the joint velocities and the end-effector linear and angular velocities. This mapping is described by a matrix, termed geometric Jacobian, which depends on the manipulator configuration.

In total, we tend to describe the end effector linear velocity $\dot{p}$ and angular velocity $\omega$ as a function of joint velocities $\dot{q}$.

$$v_{(6\times1)} = \begin{bmatrix} \dot{p} \\ \omega \end{bmatrix}_{(6\times1)} = J(q)\dot{q} = \begin{bmatrix} J_{P(3\times n)} \\ J_{O(3\times n)} \end{bmatrix}_{(6\times n)} \dot{q}_{(n\times1)}$$

The ($6\times n$) matrix J is the manipulator geometric Jacobian which in general is a function of the joint variables.

### A. Jacobian Computation

In this section we derive the Jacobian Matrices for WAM & OUR arms. Both manipulators consist of 6 revolute joints (n=6) and here we define step-by-step derivation of Jacobian matrices.

We showed that the jacobian matrice can be shown as ($6\times n$) matrices. J can be partitioned into ($3\times1$) column vectors. In these arms the J matrices are ($6\times6$). And can be shown as:

$$J = \begin{bmatrix} J_{p1} & J_{p2} & J_{p3} & J_{p4} & J_{p5} & J_{p6} \\ J_{O1} & J_{O2} & J_{O3} & J_{O4} & J_{O5} & J_{O6} \end{bmatrix}$$

Since all the joints are revolute, $J_{Pi}$ and $J_{Oi}$ are computed by:

$$J_{Pi} = z_{i-1} \times (P - p_{i-1})$$
$$J_{Oi} = z_{i-1}$$

Where P is the position of end effector with respect to Base reference, Figure(5). $p_{i-1}$ is the position of each revolute joint with respect to base frame and can be derived from the first three elements of the fourth column of the transformation matrix $T_n^0$.

$$p_{i-1} = A_1^0(q_1)...A_{i-1}^{i-2}(q_{i-1})p_0$$

Where i= 1:6 and $p_0 = \begin{bmatrix} 0 & 0 & 0 & 1 \end{bmatrix}^T$ allows selecting the fourth desired column.

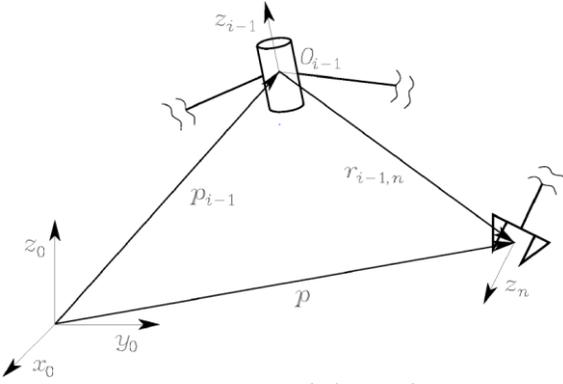

*Figure 5: Vectors neede for Jacobian computation*

$z_{i-1}$ is the joint axis vector of each revolute joint and can be expressed by:

$$z_{i-1} = R_1^0(q_1)...R_{i-1}^{i-2}(q_{i-1})z_0$$

Where i= 1:6 and $z_0 = \begin{bmatrix} 0 & 0 & 1 \end{bmatrix}^T$ allows selecting the third desired column.

## IV. KINEMATIC SINGULARITY

While singularities have non-local implications for the control and use of manipulators, they arise as local or instantaneous phenomena from the rank deficiency of a derivative. For serial manipulators, it is the singularities of the kinematic mapping/forward kinematics and trajectories that are of interest, whereas for fully parallel manipulators it is those of the constraint function defining the configuration space and of the projection onto the articular space (inverse kinematics). The distinction between the classes of mechanisms in respect of their singularities was first recognized by Gosselin and Angeles [2] and subsequently refined by Zlatanov et al in [3] Simaan and Shoham have used their ideas to analyze singularities of hybrid serial/in–parallel mechanisms [4]. The importance of singularities from an engineering perspective arises for several reasons [5]:

*1) Loss of freedom:*

The derivative of the kinematic mapping or forward kinematics represents the conversion of joint velocities into generalized end-effector velocities, i.e. linear and angular velocities. This linear transformation is generally referred to as the manipulator Jacobian in the robotics literature. A drop in rank reduces the dimension of the image, representing a loss of instantaneous motion for the end effector of one or more degrees.

*2) Workspace:*

When a manipulator is at a boundary point of its workspace, the manipulator is necessarily at a singular point of its kinematic mapping, though the converse is not the case. Interior components of the singular set separate regions with different numbers or topological types of inverse kinematics. These are usually associated with a change of posture in some component of the manipulator. Therefore knowledge of the manipulator singularities provides valuable information about its workspace [6].

*3) Loss of control:*

A variety of control systems is used for manipulators. Rate control systems require the end–effector to traverse a path at a fixed rate and therefore determine the required joint velocities by means of the inverse of the derivative of the (known) forward kinematics. Near a singularity, this matrix is ill-conditioned and either the control algorithm fails or the joint velocities and accelerations may become unsustainably great. Conversely, force control algorithms, well-adapted for parallel manipulators, may result in intolerable joint forces or torques near singularities of the projection onto the joint space.

*4) Mechanical advantage:*

Near a singular configuration, large movement of joint variables may result in small motion of the end–effector. Therefore there is mechanical advantage that may be realised as a load-bearing capacity or as fine control of the end effector. Another aspect of this is in the design of mechanisms possessing trajectories with specific singularity characteristics.

## V. MATLAB CODE EXPLANATION

### A. Transfer matrices function

A function is written in Matlab that gets the D-H parameters as Input and gets the $A_{i-1}^{i-2}$ as output:

```
%%%%%% Trans.m %%%%%%
function [ T ] = Trans( a,b,c,d )
  % D-H Homogeneous Transformation Matrix
  (a alpha d theta)
 T = [
 cos(d) -sin(d)*round(cos(b))
 sin(d)*sin(b)  a*cos(d);
 sin(d)  cos(d)*round(cos(b)) -
 cos(d)*sin(b)  a*sin(d);
 0 sin(b)  round(cos(b))  c;
 0 0 0 1
     ];
end
```

### B. Inserting D-H Parameters

In this part we insert the D-H parameters manually to the code.

```
% Inserting D-H convention parameters
% WAM
A1 = Trans(0,-pi/2,0,t1);
A2 = Trans(0.5518,0,0,t2);
A3 = Trans(-0.45,pi/2,0,t3);
A4 = Trans(0,-pi/2,0.3,t4);
A5 = Trans(0,pi/2,0,t5);
A6 = Trans(0,0,0,t6);

% Smokie OUR
% A1 = Trans(0,pi/2,0,t1);
% A2 = Trans(0.43,0,0.145,t2);
% A3 = Trans(0.336,0,-0.145,t3);
% A4 = Trans(0,-pi/2,0.115,t4);
% A5 = Trans(0,pi/2,0.115,t5);
% A6 = Trans(0,0,0.115,t6);
```

## C. Creating Transfer Matrices

After creating each transfer Matrix $A_i^{i-1}$ and by postmultiplying them, we have our Transformation matrix of $A_{ee}^0$ named as T6.

```
%Creating Transfer matrices
 T2= A1*A2;
 T3= A1*A2*A3;
 T4= A1*A2*A3*A4;
 T5= A1*A2*A3*A4*A5;
 T6= A1*A2*A3*A4*A5*A6;
```

## D. Creating $p_{i-1}$ and $z_{i-1}$ and $P$

For computing the jacobian matrices, we create $p_{i-1}$ and $z_{i-1}$ and P matrices, as expressed in section III.

```
% Creating zi

 z0= [0;0;1];
 z1= A1(1:3,3);
 z2= T2(1:3,3);
 z3= T3(1:3,3);
 z4= T4(1:3,3);
 z5= T5(1:3,3);

% Creating pi

 p0=[0;0;0];
 p1=A1(1:3,4);
 p2=T2(1:3,4);
 p3=T3(1:3,4);
 p4=T4(1:3,4);
 p5=T5(1:3,4);

 P=T6(1:3,4);
```

## E. Jacobian Matrix computation

In this part of program, we compute the jacobian Matrix. We use the simplify command to get more concise equations. In order to derive the decoupled singularities, the $(3\times 3)$ blocks' Jacobians are computed.

```
% Jacobian matrix Computation
J= simplify(
[cross(z0,P-p0),cross(z1,P-p1,
cross(z2,P-p2),cross(z3,P-p3),
cross(z4,P-p4),cross(z5,P-p5);
z0        ,   z1        ,   z2
z3        ,   z4        ,   z5
])

% (3*3) blocks Jacobians
J11=J(1:3,1:3);
J22=J(4:6,4:6);
```

## F. Jacobian Matrice Determinant

Finally the determinant of each Jacobian is calculated. Simplify command help the result to be simpler.

```
% Determinant Calculation
det00=simplify(det(J));
det11=simplify(det(J11));
det22=simplify(det(J22));
```

## VI. WAM AND OUR JACOBIAN AND SINGULARITIES

By running the MATLAB program for each of these two arms we will get the Jacobian matrices and their determinants.

### A. BARRETT WAM

When we study the 6 DoF WAM, we find out that it is approximately similar to an anthropomorphic arm and it has a spherical wrist. As singularities are typical of the mechanical structure and do not depend on the frames chosen to describe kinematics, it is convenient to choose the origin of the end effector frame at the intersection of the wrist axes. By choosing $p = p_w$ the up-right $(3\times 3)$ block will be zeo and Jacoubian Matrice can be shown as:

$$J = \begin{bmatrix} J_{11} & J_{12} \\ J_{21} & J_{22} \end{bmatrix} = \begin{bmatrix} J_{11} & 0 \\ J_{21} & J_{22} \end{bmatrix}$$

For this purpose ($p = p_w$), the last line of D-H table will change. In fact $d6 = 0$

*Table 3: Editet WAM D-H Table*

| i | $a_i$ | $\alpha_i$ | $d_i$ | $\theta_i$ |
|---|---|---|---|---|
| 1 | 0 | $-\pi/2$ | 0 | $\theta_1$ |
| 2 | $\sqrt{0.55^2 + 0.045^2}$ | 0 | 0 | $\theta_2$ |
| 3 | -0.045 | $\pi/2$ | 0 | $\theta_3$ |
| 4 | 0 | $-\pi/2$ | 0.3 | $\theta_4$ |
| 5 | 0 | $\pi/2$ | 0 | $\theta_5$ |
| 6 | 0 | 0 | 0 | $\theta_6$ |

Since all vectors $p_w - p_i$ are parallel to the unit vectors Zi, for i = 3,4,5, no matter how Frames 3, 4,5 are chosen according to Denavit-Hartenberg convention. In view of this choice, the overall Jacobian becomes a block lower triangular matrix. In this case, computation of the determinant is greatly simplified, as this is given by the product of the determinants of the two blocks on the diagonal:

$$\det(J) = \det(J_{11})\det(J_{22})$$

As a result the singularity decouping has been achieved. For determining of Arm Singularities, we use:

$$\det(J_{11}) = 0$$

And for finding wrist singularities we solve the equation of:

$$\det(J_{22}) = 0$$

*1) WAM Arm Singularity*

The MATLAB code is run for WAM arm and the jacobian matrix is derived. As it can be seen in Appendix A, the Geometric Jacobian is in accordance with singularity decoupled Jacobian introduced in above. The J is a full rank matrix.

The determinant of Up-Left $(3\times3)$ block matrice, names az det11 is computed

Det11=
0.1490*cos(t2)*cos(t3)^2 - 0.1117*sin(t2) - 0.0913*cos(t2)*cos(t3) - 0.1370*cos(t2)*sin(t3) - 0.074493*cos(t2) + 0.0621*cos(t3)^2*sin(t2) - 0.1490*cos(t3)*sin(t2)*sin(t3) + 0.0621*cos(t2)*cos(t3)*sin(t3)

By analyzing the Det(J11) we understand that the singularities of WAM arm are dependent only on $\theta_2$ and $\theta_3$.

This is a complicated equation. We try to find the angles which make the Det(J11) = 0.
First we try to use the "Solve" Command in MATLAB.
The answers which MATLAB shows, all have imaginary part.

>> solve(det11,t2)

ans =
 pi*k - (log(-(exp(t3*i)*2759*i - 1500 - 2250*i)/(exp(t3*i)*2759*i + exp(t3*2*i)*(1500 - 2250*i)))*i)/2

If we want to see the real answers, No answers are shown.

For making Sure a Code is written, which plots the determinant with "Meshgrid" command. It can be seen in figure(6) that there are points where Det(J11) is equal to zero.

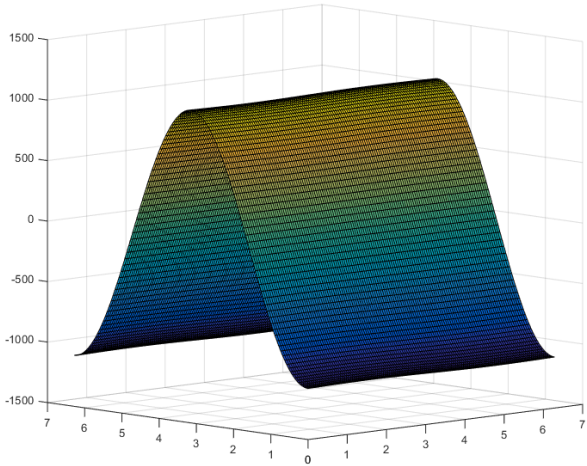

Figure 6: 3-D plot od feterminant function for WAM

To find the angles by which determinant gets zero, a MATLAB code is used. This code calculates determinant discretely, with 1° accuracy steps. The angles which make the determinant less than $10^{-6}$ (supposed to be zero) are:

Table 4: probable angles cause singularity

| $\theta_2$ | $\theta_3$ |
|---|---|
| 42 | 325 |
| 222 | 325 |
| 72-73-74-75-76-77-78 | 326 |
| 252-253-254-255-256-257-258 | 326 |
| 120-121 | 327 |
| 300-301 | 327 |
| 145 | 328 |
| 325 | 328 |

The behavior of Determinant is oscillatory based on different frequencies (Same as Signals). The figure (7) shows the det(J11).

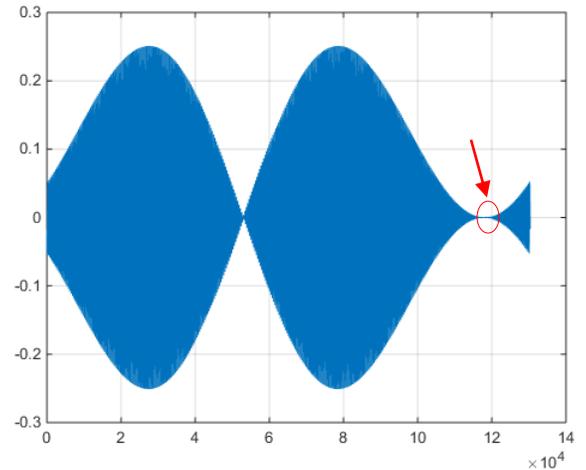

Figure 7: Determinant behaviour

If we narrow our focus to $\theta_3 = 324$ - $\theta_3 = 329$ on figure(8), as shown by red circle we conclude that a singularity exists in this range This angle is approximately $\theta_3 = 326°$

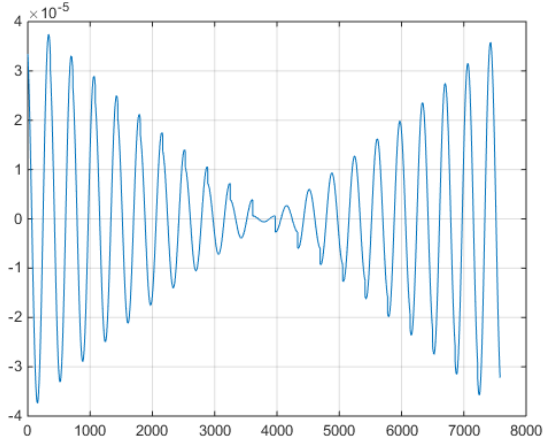

Figure 8: Narrowed view of probable singularity point

2) *Wrist Singularities*

For finding the wrist singularities the $\det(J_{22}) = 0$ equation is calculated.

$$det22 = -\sin(t5) = 0$$

So, The Wrist singularities: $\theta_5 = 0, \pi$.

## VII. OUR JACOBIANS AND SINGULARITY

The MATLAB code is followed for OUR robot. Since this Arm doesn't have the spherical wrist, we have to compute the total Jacobian and derive its determinant. The jacobian matrix is shown in Appendix B. it's a full rank matrix.
By using the "det" command in MATLAB the determinant is calculated. The simplified det(J) is:

0.00014448*sin(t5)*(168.0*sin(t2 + 2.0*t3) - 57.5*cos(t2 + t4) + 215.0*sin(t2 + t3) + 57.5*cos(t2 + 2.0*t3 + t4) - 215.0*sin(t2 - 1.0*t3) - 168.0*sin(t2))

As it's obvious this determinant is a function of $\theta_2, \theta_3, \theta_4$ and $\theta_5$. By solving the equation $\det(J) = 0$, the singularities are found. The first two sets of singularities are:

$$\sin(\theta_5) = 0 \Rightarrow \theta_5 = 0, \pi$$

And

$$\sin(\theta_3) = 0 \Rightarrow \theta_3 = 0, \pi$$

By plotting the above multivariable determinant function, based on $\theta_3$ (the dominant frequency) we see that there are a great number of points which makes the determinant zero. It's obvious in figure(9) that at $\theta_3 = 0, \pi$, we have Singularity.

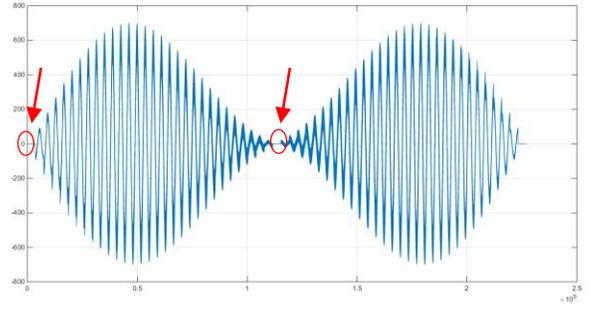

Figure 9: Determinant multivariable function behavior

But for other points which intersect with zero line, I don't think that they are singularities. Because so many singularities make the Arm useless, since there exists a lot of situation which the manipulation should avoid them.
When we try to solve this equation with "solve" command in MATLAB, the results always contains imaginary part and the equation doesn't have pure real answer. I think these intersections would not happen in reality and they are not singularity.
Also the OUR design is based on UR5 arm robot. They have very similar design and properties. No such problem about multiple singularities discussed in literature yet.
So, I conclude that they are not singularities

## VIII. SINGULARITY CONFIGURATIONS

### A. WAM singularity configurations

WAM singularities are shown in below figures.
The ($\theta_5 = \pi$) singularity is shown schematic.

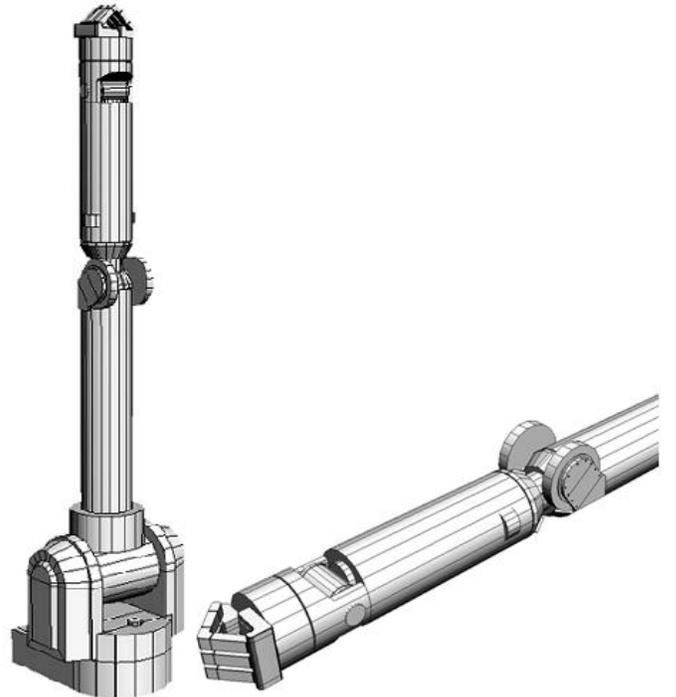

Figure 10: Wrist singularity ($\theta_5 = 0$) and Arm Singularity ($\theta_3 = 0$)

## B. OUR singularity configuration

In the below figures, the OUR singularities are shown.

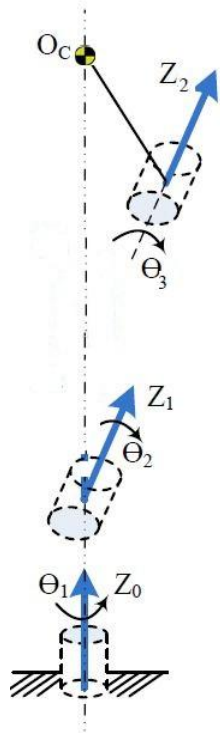

*Figure 12: Shoulder Singularity*

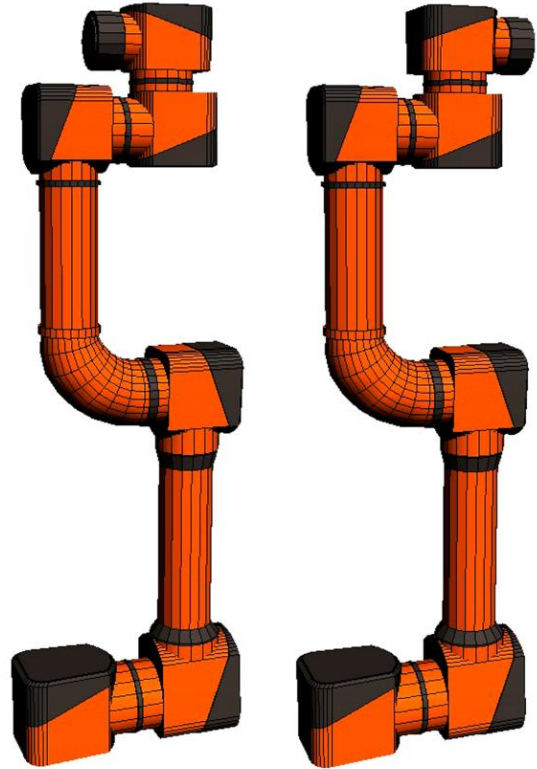

*Figure 13: OUR Singularity $\theta_5 = 0$*

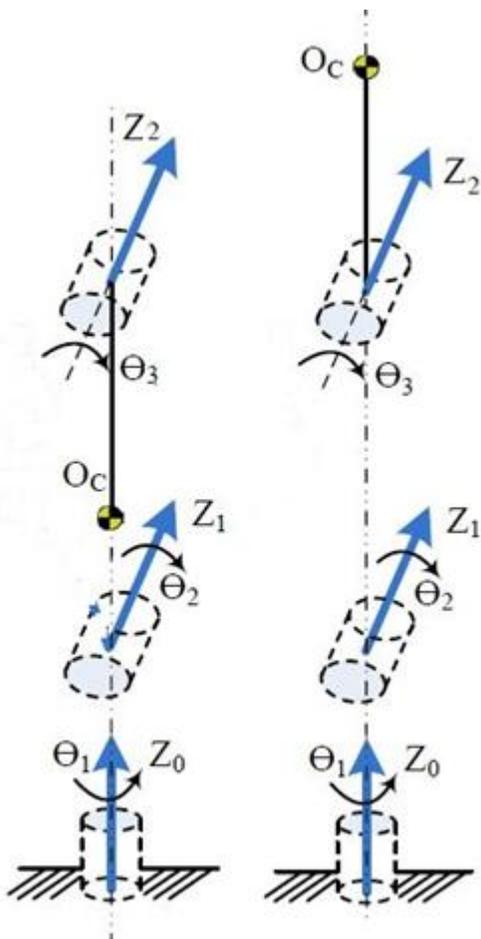

*Figure 11: Arm Singularities*

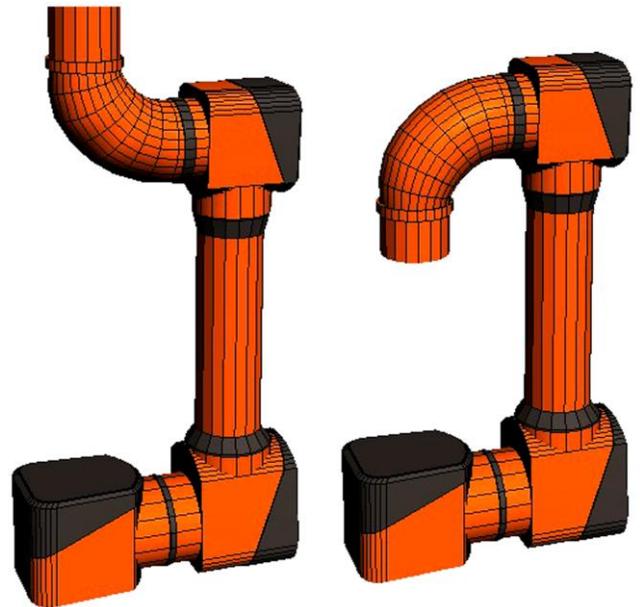

*Figure 14: OUR Singularity $\theta_3 = 0$*

## IX. REDUNDANT KINEMATIC ALLOCATION SCHEMES FOR WAM

The robot has more than 6 DOF then it is termed redundant and there may be many inverse kinematic (IK) solutions for a given end-effector configuration. Often, 7-DoF redundant manipulators IK problems are solved iteratively with methods that rely on the inverse Jacobian pseudoinverse Jacobian or Jacobian transpose. These approaches are generally slow and sometimes suffer from singularity issues.

The Redundancy Allocation is to select the optimal combination of components and redundancy degree to meet resource constraints while maximizing the system reliability.

In Robotics In order to make the robot more dexterous for unpredictable and varying environment, researchers have to study into the redundant problem of a robot with more than 6 DoFs. Barrett WAM has been chosen for that study by many scholars.

Giresh K. Singh and Jonathan Claassens [7] An analytical solution to the inverse kinematics problem for the 7 Degrees of Freedom Barrett Whole Arm Manipulator with link offsets. A method to obtain all possible geometric poses (both the in-elbow & out-elbow) for a desired end-effector position and orientation is provided. The set of geometric poses is completely determined by three circles in the Cartesian space. The joint-variables can be easily computed for any geometric pose. The physical constraints on the joint-angles restrict the set of feasible poses. The constraints on the set of feasible poses have been analytically worked out for the joint-variables of the 'cosine' type.

H.Y.K. Lau and L.C.C.Wai [8] studied the redundant control strategy of the 7-DoF WAM. Choosing the third joint as the redundant DoF, they tackled the redundant kinematics problem for WAM by the Jacobian method. Compared with the analytical method, Jacobian method is less computational economical. Jacobian method advantage is its fast track for formulating a serial inverse kinematic problems, such as WAM. Jacobian method is flexible to allow most optimization algorithm to build upon. This Redundant control strategy mechanisms constructed to effectively switch the appropriate algorithm that is most suitable to control the arm under different situations. This control strategy use the basis of computer-based feedback controller.

## X. CONCLUSION

In this paper the jacobian computation and Singularity identification are expressed thoroughly and the MATLAB code is run to show the result for two 6 DoF arm robot, BARRETT WAM and SMOKIE OUR. After deriving the determinants, a detailed discussion is done to find the singularities. Different approaches are used to find the singularity point and regions. The Singularity configurations for these two robots are shown by CAD 3D design and schematically. Finally we discuss about redundant kinematic allocation for WAM robot.

# XII. APPENDIX

## A. Jacobian Matrix of WAM

J =

[ -(sin(t1)*(1500*sin(t2 + t3) - 2250*cos(t2 + t3) + 2759*cos(t2)))/5000, cos(t1)*((3*cos(t2 + t3))/10 + (9*sin(t2 + t3))/20 - (2759*sin(t2))/5000), cos(t1)*((3*cos(t2 + t3))/10 + (9*sin(t2 + t3))/20), 0, 0, 0]

[ (cos(t1)*(1500*sin(t2 + t3) - 2250*cos(t2 + t3) + 2759*cos(t2)))/5000, sin(t1)*((3*cos(t2 + t3))/10 + (9*sin(t2 + t3))/20 - (2759*sin(t2))/5000), sin(t1)*((3*cos(t2 + t3))/10 + (9*sin(t2 + t3))/20), 0, 0, 0]

[ 0, (9*cos(t2 + t3))/20 - (3*sin(t2 + t3))/10 - (2759*cos(t2))/5000, (9*cos(t2 + t3))/20 - (3*sin(t2 + t3))/10, 0, 0, 0]

[ 0, -sin(t1), -sin(t1), sin(t2 + t3)*cos(t1), sin(t4)*(cos(t1)*sin(t2)*sin(t3) - cos(t1)*cos(t2)*cos(t3)) - cos(t4)*sin(t1), cos(t5)*(cos(t1)*cos(t2)*sin(t3) + cos(t1)*cos(t3)*sin(t2)) - sin(t5)*(sin(t1)*sin(t4) + cos(t4)*(cos(t1)*sin(t2)*sin(t3) - cos(t1)*cos(t2)*cos(t3)))]

[ 0, cos(t1), cos(t1), sin(t2 + t3)*sin(t1), cos(t1)*cos(t4) + sin(t4)*(sin(t1)*sin(t2)*sin(t3) - cos(t2)*cos(t3)*sin(t1)), sin(t5)*(cos(t1)*sin(t4) - cos(t4)*(sin(t1)*sin(t2)*sin(t3) - cos(t2)*cos(t3)*sin(t1))) + cos(t5)*(cos(t2)*sin(t1)*sin(t3) + cos(t3)*sin(t1)*sin(t2))]

[ 1, 0, 0, cos(t2 + t3), sin(t2 + t3)*sin(t4), cos(t2 + t3)*cos(t5) - sin(t2 + t3)*cos(t4)*sin(t5)]

## B. Jacobian matrix of OUR

J =

[ (23*cos(t1))/200 + (23*cos(t1)*cos(t5))/200 - (43*cos(t2)*sin(t1))/100 + (42*sin(t1)*sin(t2)*sin(t3))/125 - (23*cos(t2 + t3 + t4)*sin(t1)*sin(t5))/200 + (23*cos(t2 + t3)*sin(t1)*sin(t4))/200 + (23*sin(t2 + t3)*cos(t4)*sin(t1))/200 - (42*cos(t2)*cos(t3)*sin(t1))/125, -cos(t1)*((42*sin(t2 + t3))/125 + (43*sin(t2))/100 - (23*sin(t2 + t3)*sin(t4))/200 + sin(t5)*((23*cos(t2 + t3)*sin(t4))/200 + (23*sin(t2 + t3)*cos(t4))/200) + (23*cos(t2 + t3)*cos(t4))/200), -cos(t1)*((23*cos(t2 + t3 + t4))/200 + (42*sin(t2 + t3))/125 + (23*sin(t2 + t3 + t4)*sin(t5))/200), -cos(t1)*((23*cos(t2 + t3 + t4))/200 + (23*sin(t2 + t3 + t4)*sin(t5))/200), (23*cos(t1)*cos(t2)*cos(t3)*cos(t4)*cos(t5))/200 - (23*sin(t1)*sin(t5))/200 - (23*cos(t1)*cos(t2)*cos(t5)*sin(t3)*sin(t4))/200 - (23*cos(t1)*cos(t3)*cos(t5)*sin(t2)*sin(t4))/200 - (23*cos(t1)*cos(t4)*cos(t5)*sin(t2)*sin(t3))/200, 0]

[ (23*sin(t1))/200 + (43*cos(t1)*cos(t2))/100 + (23*cos(t5)*sin(t1))/200 - (42*cos(t1)*sin(t2)*sin(t3))/125 + (23*cos(t2 + t3 + t4)*cos(t1)*sin(t5))/200 - (23*cos(t2 + t3)*cos(t1)*sin(t4))/200 - (23*sin(t2 + t3)*cos(t1)*cos(t4))/200 + (42*cos(t1)*cos(t2)*cos(t3))/125, -sin(t1)*((42*sin(t2 + t3))/125 + (43*sin(t2))/100 - (23*sin(t2 + t3)*sin(t4))/200 + sin(t5)*((23*cos(t2 + t3)*sin(t4))/200 + (23*sin(t2 + t3)*cos(t4))/200) + (23*cos(t2 + t3)*cos(t4))/200), -sin(t1)*((23*cos(t2 + t3 + t4))/200 + (42*sin(t2 + t3))/125 + (23*sin(t2 + t3 + t4)*sin(t5))/200), -sin(t1)*((23*cos(t2 + t3 + t4))/200 + (23*sin(t2 + t3 + t4)*sin(t5))/200), (23*cos(t1)*sin(t5))/200 + (23*cos(t2)*cos(t3)*cos(t4)*cos(t5)*sin(t1))/200 - (23*cos(t2)*cos(t5)*sin(t1)*sin(t3)*sin(t4))/200 - (23*cos(t3)*cos(t5)*sin(t1)*sin(t2)*sin(t4))/200 - (23*cos(t4)*cos(t5)*sin(t1)*sin(t2)*sin(t3))/200, 0]

[ 0, (23*sin(t2 + t3 + t4 + t5))/400 - (23*sin(t2 + t3 + t4))/200 - (23*sin(t2 + t3 + t4 - t5))/400 + (42*cos(t2 + t3))/125 + (43*cos(t2))/100, (23*sin(t2 + t3 + t4 + t5))/400 - (23*sin(t2 + t3 + t4))/200 - (23*sin(t2 + t3 + t4 - t5))/400 + (42*cos(t2 + t3))/125, (23*sin(t2 + t3 + t4 + t5))/400 - (23*sin(t2 + t3 + t4))/200 - (23*sin(t2 + t3 + t4 - t5))/400, (23*sin(t2 + t3 + t4 - t5))/400 + (23*sin(t2 + t3 + t4 + t5))/400, 0]

[ 0, sin(t1), sin(t1), sin(t1), -sin(t2 + t3 + t4)*cos(t1), cos(t5)*sin(t1) + cos(t2 + t3 + t4)*cos(t1)*sin(t5)]

[ 0, -cos(t1), -cos(t1), -cos(t1), -sin(t2 + t3 + t4)*sin(t1), cos(t2 + t3 + t4)*sin(t1)*sin(t5) - cos(t1)*cos(t5)]

[ 1, 0, 0, 0, cos(t2 + t3 + t4), sin(t2 + t3 + t4)*sin(t5)]